\documentclass{article}

\usepackage[preprint]{neurips_2024}

\usepackage[utf8]{inputenc} 
\usepackage[T1]{fontenc}    
\usepackage{hyperref}       
\usepackage{url}            
\usepackage{booktabs}       
\usepackage{amsfonts}       
\usepackage{nicefrac}       
\usepackage{microtype}      
\usepackage{xcolor}         

\usepackage{natbib} 
\setcitestyle{numbers,square}
\usepackage{graphicx}
\usepackage{multirow}
\newlength{\Oldarrayrulewidth}
\newcommand{\Cline}[2]{%
  \noalign{\global\setlength{\Oldarrayrulewidth}{\arrayrulewidth}}%
  \noalign{\global\setlength{\arrayrulewidth}{#1}}\cline{#2}%
  \noalign{\global\setlength{\arrayrulewidth}{\Oldarrayrulewidth}}}
\usepackage{pifont}
\usepackage{amsmath}

\title{Learn To Learn More Precisely}

\author{
    \textbf{Runxi Cheng\textsuperscript{1}, Yongxian Wei\textsuperscript{1}, Xianglong He\textsuperscript{1}, Wanyun Zhu\textsuperscript{2}} \\
    \textbf{Songsong Huang\textsuperscript{3}, Fei Richard Yu\textsuperscript{4}, Fei Ma\textsuperscript{4}, Chun Yuan\textsuperscript{1,$\dagger$}} \\
    \textsuperscript{1}Tsinghua Shenzhen International Graduate School, Tsinghua University; \textsuperscript{2}CUHK-Shenzhen; \\
    \textsuperscript{3}Fudan University;\textsuperscript{3}Guangdong Laboratory of Artificial Intelligence and Digital Economy (SZ)\\
	\textsuperscript{*}\href{mailto:crx23@mails.tsinghua.edu.cn}{\texttt{crx23@mails.tsinghua.edu.cn}}; \textsuperscript{$\dagger$}\href{mailto:yuanc@sz.tsinghua.edu.cn}{\texttt{yuanc@sz.tsinghua.edu.cn}}
}

\begin{document}

\maketitle

\begin{abstract}

Meta-learning has been extensively applied in the domains of few-shot learning and fast adaptation, achieving remarkable performance. While Meta-learning methods like Model-Agnostic Meta-Learning (MAML) and its variants provide a good set of initial parameters for the model, the model still tends to learn shortcut features, which leads to poor generalization. In this paper, we propose the formal conception of ``learn to learn more precisely'', which aims to make the model learn precise target knowledge from  data and reduce the effect of noisy knowledge, such as background and noise. To achieve this target, we proposed a simple and effective meta-learning framework named Meta Self-Distillation(MSD) to maximize the consistency of learned knowledge, enhancing the models' ability to learn precise target knowledge. In the inner loop, MSD uses different augmented views of the same support data to update the model respectively. Then in the outer loop, MSD utilizes the same query data to optimize the consistency of learned knowledge, enhancing the model's ability to learn more precisely. Our experiment demonstrates that MSD exhibits remarkable performance in few-shot classification tasks in both standard and augmented scenarios, effectively boosting the accuracy and consistency of knowledge learned by the model.

\end{abstract}

\section{Introduction}

Meta-learning, also known as ``learning to learn'', aims to endow models with rapid learning capabilities\cite{lake2023human,finn2017model}. This includes the ability to recognize objects from a few examples or to quickly acquire new skills after minimal exposure. Meta-learning can be broadly categorized into two main factions: metric-based\cite{snell2017prototypical,zhou2023revisiting,vinyals2016matching,sung2018learning,sun2020meta} and optimize-based meta-learning\cite{finn2017model,antoniou2018train,ye2021train,jamal2019task}. Metric-based approaches typically strive to learn generalized features that perform well on tasks such as few-shot learning. However, during the meta-testing phase, metric-based meta-learning does not usually involve fine-tuning, implying that the model learns more general features rather than learning how to learn. Optimization-based meta-learning, on the other hand, is primarily considered to embody the concept of learning to learn. The mainstream method in this domain is Model-Agnostic Meta-Learning (MAML)~\cite{finn2017model} and its numerous variants~\cite{raghu2019rapid,ye2021train,antoniou2018train,kao2021maml,nichol2018first}. Despite MAML and its derivatives demonstrating impressive performance on few-shot tasks, they still exhibit certain flaws. Current research~\cite{le2021poodle,zhou2023revisiting} indicates that models tend to learn shortcut features(e.g., color, background, etc.) that are exclusively sufficient to distinguish very few classes in the meta-training phase. Models trained with MAML inevitably encounter these issues, and most improvements to MAML focus on optimizing the method without addressing the bias in model learning. Poodle~\cite{le2021poodle} has proposed a regularization approach using additional data, but this does not enhance the model's ability to learn invariant features in the few-shot scenario. Therefore, the challenge of how to learn more precisely remains a critical problem in meta-learning.

\begin{figure}[tbp]
	\centering
	\includegraphics[width=1\textwidth]{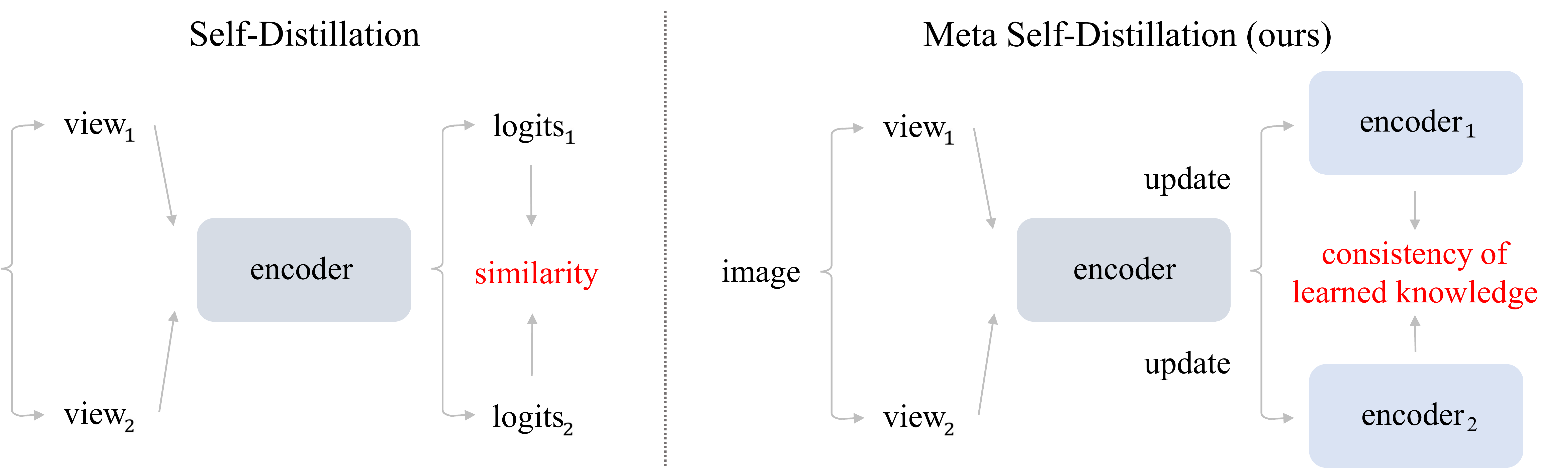}
	\caption{\textbf{The core idea between Self-Distillation and Meta Self-Distillation.} Self-Distillation aims to make the deep representation of different views closer, while Meta Self-Distillation aims to learn the same knowledge from the different views of the same image.}
\end{figure}

In this paper, we propose the concept of ``learn to learn more precisely''. Our goal is to enable models to learn more accurate knowledge from training data. In few-shot tasks, due to the scarcity of samples, models may consider noise and background in the input as the core features related to classification, inevitably leading to overfitting. Drawing on the knowledge concept introduced by~\cite{hinton2015distilling}, we propose the formal concept of  `` knowledge'' and ``the change of knowledge''. Based on this, we further defined the problem target of ``learn to learn more precisely''. We propose that models should learn the accurate target knowledge instead of the noisy knowledge, which means when a model learns knowledge from a certain image with different noises, ``the change of knowledge'' should be the same.

From the perspective of learn to learn more precisely, we propose Meta Self-Distillation (MSD), a simple and effective meta-learning framework for few-shot classification. Specifically, given random mini-batch data, we update the initial parameter with different augmented views of the data respectively in the inner loop, then use the same query data to measure the consistency of learned knowledge among each updated model in the outer loop. By maximizing the consistency in outputs for the same query data across different updated models, we enhance the initial parameters' ability to learn more precisely. Throughout extensive experiments, we demonstrate the effectiveness of MSD. In both standard and augmented few-shot classification problems, MSD outperforms many recent few-shot classification algorithms, and in the augmented scenario for few-shot tasks, MSD achieved an average improvement of 7.42\% and 4.03\% in the 5way1shot and 5way5shot problems, respectively. Also, we define to use cosine similarity between the predictions and mean predictions to measure the consistency of learned knowledge, and we achieve remarkable performance in this metric

In summary, the contributions of our work are threefold:
\begin{itemize}
\item Based on the knowledge concept proposed by~\cite{hinton2015distilling}, we introduce the formal notion of ``knowledge'' and ``the change of knowledge'' and further propose the optimization goal of ``learn to learn precisely'', which aims to learn precise target knowledge and reduce the effect of noisy knowledge.
\item Drawing inspiration from MAML and self-distillation, we propose Meta Self-distillation to enable the initial parameters to have the ability to learn more precisely. By maximizing the consistency of knowledge learned from different views of the same instance, the model gains the ability to learn more precisely.
\item Our experiments demonstrate that our method effectively enhances the model's performance in few-shot learning and exhibits a superior ability to learn precisely in more complex few-shot classification problems compared to existing meta-learning algorithms.
\label{intro}
\end{itemize}

\section{Related Work}
\subsection{Meta-Learning}

Meta-learning, also known as learning to learn, endows models with strong learning capabilities after the meta-training phase. It is mainly divided into metric-based meta-learning, represented by ProtoNet~\cite{snell2017prototypical}, and optimization-based meta-learning, represented by MAML\cite{finn2017model}. Metric-based meta-learning improves model representation by bringing the deep representation between the support data and the query data that belong to the same category closer, typically not requiring fine-tuning during the meta-test phase. Optimize-based meta-learning aims to provide the model with good initial weights, offering better generalization performance when fine-tuning on novel category samples. This category includes algorithms like MAML~\cite{finn2017model} and its variants, such as~\cite{ye2021train}, which utilizes a single vector to replace the network's classification head weight, thus preventing the permutation in the meta-test phase. MAML++~\cite{antoniou2018train} enhances MAML's performance by addressing multiple optimization issues encountered by MAML, while ANIL~\cite{raghu2019rapid} improves MAML's performance by freezing the backbone during the inner loop. However, these algorithms mainly improve MAML from an optimization perspective, while they do not further advance MAML in terms of increasing the model’s learning capabilities. Recent studies also suggest that meta-learning is more concerned with learning more general features from the training dataset.~\cite{raghu2019rapid} demonstrate that the effectiveness of MAML is attributed to feature reuse rather than rapid learning.~\cite{ni2021close} discuss the close relationship between contrastive learning and meta-learning under a certain task distribution. ~\cite{kao2021maml} argues that MAML is essentially noisy, supervised contrastive learning. These studies indicate that existing meta-learning algorithms may focus more on obtaining more generalized representations from the training dataset rather than on how to enhance the model's learning capabilities, while our work focuses primarily on how to enhance the model's learning capabilities to learn more precisely.

\subsection{Self-Distillation}
Self-distillation is a variant of contrastive learning\cite{caron2020unsupervised,chen2021exploring,chen2020simple,he2020momentum,li2022metaug} that trains the model by bringing the representations of different views of the same image closer without negative pairs. BYOL~\cite{grill2020bootstrap} first proposed contrastive learning without negative samples, i.e., self-distillation. SimSiam~\cite{chen2021exploring} further explored how self-distillation avoids collapse in a self-supervised setting, and~\cite{allen2020towards} suggests that self-distillation can serve as an implicit ensemble distillation, allowing the model to distinguish more view features. Self-distillation is an excellent method to enhance the model's feature extraction capabilities and can be effectively combined with meta-learning\cite{li2022metaug}. Typically, self-distillation aims to maximize the similarity of the representations across different views of the same data, while in our proposed method, we enhance the model's ability to learn precisely by maximizing the consistency of output from different updated models for the same image.

\section{Problem Definition For Learn To Learn More Precisely}
\subsection{Preliminary On MAML For Few-shot Learning}
\subsubsection{Few-shot Learning}

Few-shot learning aims to enable a model to achieve remarkable classification performance in novel classes after the meta-training phase by learning from only a small subset of samples from these new classes. Following~\cite{vinyals2016matching,chen2019closer,wang2020generalizing}, We define the few-shot classification problem as an N-way K-shot task, where there are N classes, each containing K-labeled support samples. Typically, K is small, such as 1 or 5. The data used to attempt to update the model is defined as the support set \(S(x,y)\), where each x represents the model's input, and y denotes the corresponding label for \(x\), with \(y\) belonging to the set \([1,n]\). The data used to evaluate the effectiveness of the model updates is defined as the query set \(Q\), which has a compositional structure and class inclusion consistent with the support set, but the samples contained in the query set are completely orthogonal to those in the support set. In the meta-testing phase, multiple tasks are samples from the novel classes to assess the model's learning ability by averaging the accuracy of each task. The novel classes do not overlap with the base class categories, and the entire base class dataset typically contains more data than the novel classes.

\subsubsection{Model-Agnostic Meta-Learning (MAML)}

The objective of MAML\cite{finn2017model} is to learn an initial parameter set \(\Phi\), such that when presented with a random task and a specified loss function, it can achieve a lower loss after \(k\) steps of optimization. This can be formally expressed as:

\begin{equation}
_{\phi} \mathbb{E} \left [ L(U^{k} (\phi)) \right ]
\end{equation}

where \(U^{k}\) denotes \(k\) updates of the parameter \(\phi\) using tasks sampled from the task distribution, which corresponds to adding a sequence of gradient vectors to the initial vector:

\begin{equation}
U(\phi) = \phi + g_1 + g_2 + \cdots + g_k.
\end{equation}

In few-shot learning scenarios, optimization algorithms like Adam or SGD are typically employed to update parameters. Typically, MAML utilizes a dataset \(S\) for updating parameters within \(U\), a process also referred to as the inner loop. Subsequently, a separate dataset \(Q\) is used to evaluate \(L\), with \(L\) directly updating the original parameters \(\phi\), a step known as the outer loop. The outer loop commonly employs SGD for updates, and its gradient can be computed as follows:

\begin{equation}
  \begin{aligned}
    g_{\text{MAML}} &= \frac{\partial}{\partial \phi} L (U(\phi))  \\
    &= U'(\phi)L' (\tilde{\phi}), \text{ where } \tilde{\phi} = U(\phi) \\
  \end{aligned}
\end{equation}

In Equation (2), \(U'(\phi)\) is the Jacobian matrix of the update operation \(U\). 
When optimized with Adam, the gradients are also rescaled element-wise, but this does not alter the conclusions. First-order MAML considers these gradients as constants, thereby replacing the Jacobian \(U'(\phi)\) with the identity operation.

\begin{figure}[tbp]
	\centering
	\includegraphics[width=1\textwidth]{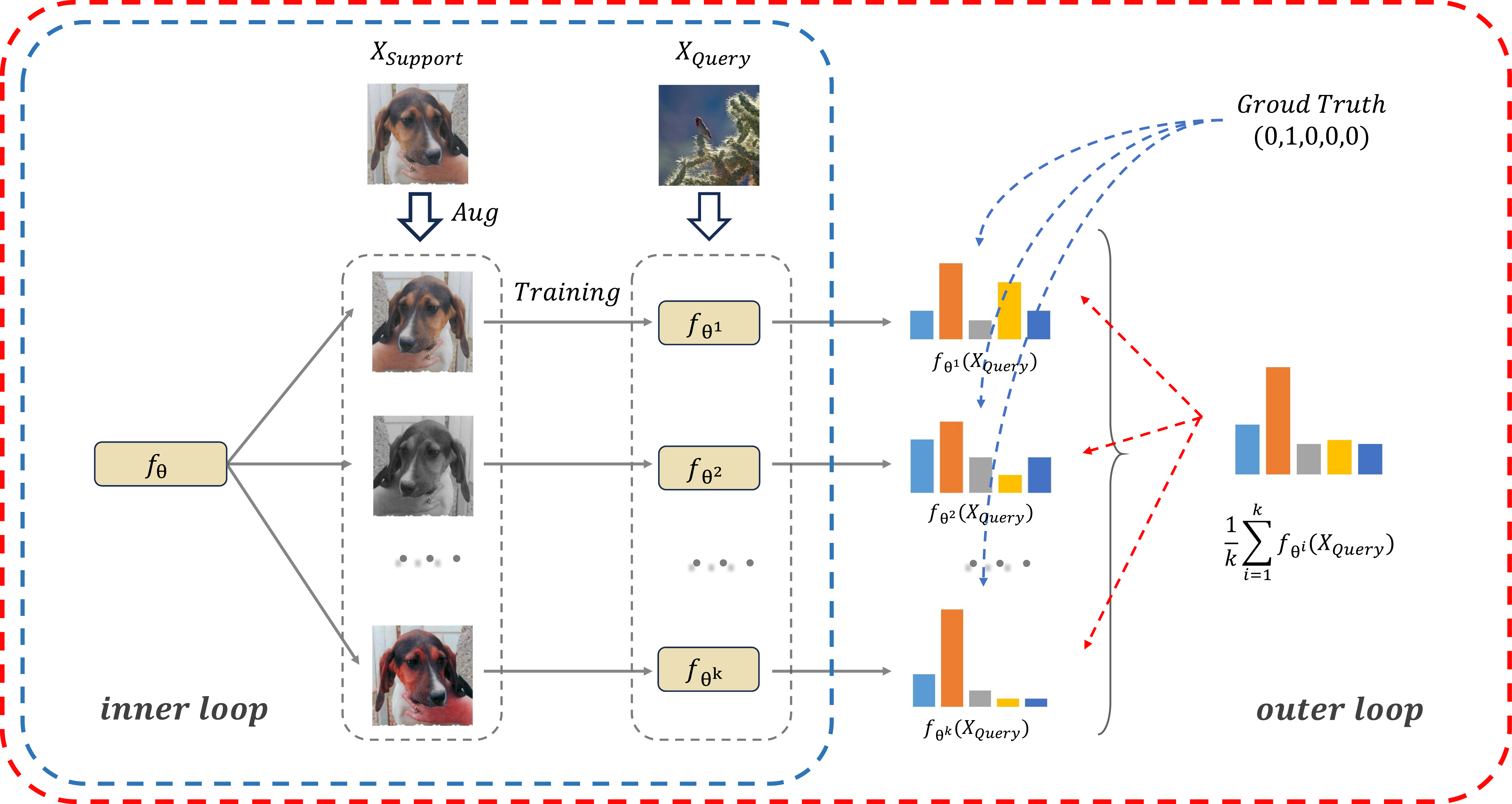}
	\caption{\textbf{An overview of the proposed MSD.} In the inner loop, MSD first uses different augmented support data to update the \(f_{\theta}\). In the outer loop, then maximizes the consistency among the outputs of the same query data with different update versions of the initial model}
\end{figure}

\subsection{Problem Definition For Learn To Learn More Precisely}
\label{Learn to learn more precisely}

MAML represents a promising learning paradigm for acquiring generalization capabilities. However, during the fine-tuning phase, due to the limited number of samples in few-shot classification, the model still tends to learn shortcut features and multiple reasonable hypotheses can lead to ambiguous classifications. Therefore, we propose the concept of 'Learning to Learn More Precisely,' enabling the model to acquire more precise target knowledge from data.

First, we need to provide a formal definition of knowledge. We Followed Hinton’s definition\cite{hinton2015distilling}, that knowledge is a learned mapping from input vectors to output vectors. For a given parameter set \(\theta\), the corresponding knowledge is denoted as \(f_{\theta}\), which can be expressed as:

\begin{equation}
f_{\theta} :  x \mapsto  y  \quad x\in X \quad \text{where} \quad y = f_{\theta}(x)
\end{equation}

where \(x\) represents the input vector, \(y\) represents the corresponding output vector, and \(X\) denotes the domain of inputs. Expanding on this, for two parameter sets \(\theta_1\) and \(\theta_2\), we define the knowledge change \(\Delta k\) with respect to \(\theta_2\) relative to \(\theta_1\) as the mapping of the change in input vectors to the change in output vectors, expressed as:

\begin{equation}
\Delta k(\theta_1,\theta_2):  x \mapsto \Delta y  \quad x\in X \quad \text{where} \quad \Delta y = \Delta k(\theta_1,\theta_2)(x)= f_{\theta_2}(x) - f_{\theta_1}(x)
\end{equation}

where \(f_{\theta_2}\) and \(f_{\theta_1}\) represent the knowledge corresponding to \(\theta_2\) and \(\theta_1\), respectively.
We assume that the knowledge of data contains the target knowledge relevant to the core features for classification and the noisy knowledge. Assuming \(\theta\) is the initial parameters and \(\theta'\) represents the parameters post-learning, the acquired knowledge can be divided into two components:

\begin{equation}
\label{eq6}
\Delta k(\theta,\theta') =  \Delta k(\theta,\theta')_{\text{target}} + \Delta k(\theta,\theta')_{\text{noise}}
\end{equation}

\(\Delta k(\theta,\theta')_{\text{target}}\) signifies the target knowledge, while \(\Delta k(\theta,\theta')_{\text{noise}}\) denotes the noisy knowledge. The corresponding changes in output are \(\Delta y_{\text{target}}\) and \(\Delta y_{\text{noise}}\), respectively. Thus, \(\Delta k(\theta,\theta')\) can be reformulated as:

\begin{equation}
\Delta k(\theta,\theta') = x \mapsto \Delta y_{\text{target}} + \Delta y_{\text{noise}}  \quad x\in X 
\end{equation}

The objective of precise learning is to minimize the influence of noisy knowledge. This objective can be defined as:

\begin{equation}
\label{eq8}
\min \int_{x\in X} \left | \Delta y_{\text{noise}}  \right | dx = \int_{x\in X} \left | \Delta k(\theta ,\theta')_{\text{noise}}(x)  \right | dx
\end{equation}

The goal of "learning to learn more precisely" is to learn a set of more "intelligent" initial parameters \(\theta\) for the model so that when faced with any training data for a specified task, the model can learn more precise target knowledge and reduce the effect of noisy knowledge. The goal can be formulated as:

\begin{equation}
\label{eq9}
\arg\min_{\theta} \int\int \left | \Delta k(\theta ,\theta')_{\text{noise}}(x)  \right | dx d\theta' = \int\int \left | \Delta y_{\text{noise}}   \right | dx d\theta'
\end{equation}

\section{Meta Self-Distillation}

Base on the target of ``learn to learn more precisely'' introduced in section \ref{Learn to learn more precisely}, we aspire for models to acquire more precise knowledge from images. As indicated in Eq.\ref{eq9}, learning precisely aims to disregard the noise knowledge in training data. For different augmented views and noise-induced variants of the same image data, we assume that the target knowledge of the images is the same. Therefore, we train the model with different augmented views of images to get the variants of the model. Then by maximizing the consistency of the variants’ output of the same query data, we can enhance the ability of the initial parameter to learn more precisely. To this end, we propose Meta Self-Distillation. This approach enables the model to learn consistent knowledge from different views of the same image, which can be calculated through the outputs of the query data.

Specifically, we sample tasks from a distribution to obtain support and query data. Unlike traditional meta-learning, which samples multiple tasks, we sample a single task and create multiple augmented views as substitutes. Augmented tasks only augmented the support data, and all augmented tasks share the same query data. The rationale behind this is to have the same standard when assessing the knowledge learned by the model.
Let a task be denoted as \( T = (S, Q) \sim P(T) \), and the task set as \( \{S(i), Q\} \), where \(S(i)\) represent the i-th .
After sampling, in the inner loop, we update the model with different augmented views of the data to obtain varied models:
\begin{equation}
\theta_i = U(\theta, S(i))
\end{equation}
In the outer loop, we test the query with different updated versions of the parameters. Since we desire the model to extract the same knowledge from different augmented views of support data, we measure the consistency of their query outputs to assess if the knowledge learned is identical:
\begin{equation}
{\mathcal L_{\text{Knowledge-Consistency}}} = \frac{1}{n} \sum f_{sim}\left(f_{\theta_i}(x_{Query}), \frac{1}{n} \sum\left(f_{\theta_i}(x_{Query})\right)\right) 
\end{equation}
Here, \( f_{sim} \) represents the function measuring output similarity. Here, we use cosine similarity.We also proof that when use consine similarity as \( f_{sim} \), out method minimize the upper bound of the modulus of noisy knowledge which was defined in Eq.\ref{eq6},which enables the model have the ability to learn precisely.

Furthermore, to ensure the model fully utilizes label information and learns precise classification, we compute the classification loss for each updated parameter by query data:
\begin{equation}
{\mathcal L_{cls}} =  \frac{1}{n}\sum   \left( {\mathcal L_{ce}}\left(f_{\theta_i}(x_{Query}), y_{Query}\right)\right) 
\end{equation}
\({\mathcal L_{ce}}\) denotes the cross-entropy loss function.
The model's total loss is expressed as:
\begin{equation}
{\mathcal L_{total}} = {\mathcal L_{\text{Knowledge-Consistency}}} + \alpha\cdot{\mathcal L_{cls}}. 
\end{equation}
where \(\alpha\) represents the coefficient of classification loss. The update in the outer loop is computed as:
\begin{equation}
\theta' = \theta - \beta \cdot {\mathcal L_{total}}(\theta, S, Q)
\label{total loss}
\end{equation}
where \(\beta\) represents the learning rate in the outer loop.

\begin{table}[htbp]
  \centering
  \small
  \caption{\textbf{5way1shot and 5way5shot classification accuracy} in standard few-shot classification task and 95\% confidence interval on {\it Mini}ImageNet and {\it Tiered}ImageNet (over 2000 tasks), using ResNet-12 as the backbone.
  }
  \label{standard_fsl}
  \vspace{-10pt}
  \tabcolsep 3pt
    \begin{tabular}{|l|c|c|cc|cc|}
    \addlinespace
    \toprule
    \multicolumn{ 3}{|c}{} & \multicolumn{ 2}{|c}{{\it Mini}ImageNet} & \multicolumn{ 2}{|c|}{{\it Tiered}ImageNet} \\
    \midrule
    \multicolumn{ 1}{|l|}{Methods} &\multicolumn{ 1}{|c|}{Backbone} &\multicolumn{ 1}{|c|}{Venue} &1-Shot & 5-Shot & 1-Shot & 5-Shot \\
    \midrule
    ProtoNet~{\cite{snell2017prototypical}}  &ResNet-12&NeurIPS'17& 62.39 \(\pm\) 0.20 & 80.53 \(\pm\) 0.20 & 68.23 \(\pm\) 0.23 & 84.03 \(\pm\) 0.16 \\
    MAML~\cite{finn2017model}  &ResNet-12&ICML'17& {64.42 \(\pm\) 0.20} & {83.44 \(\pm\) 0.14} & {65.72 \(\pm\) 0.20} & {84.37 \(\pm\) 0.16} \\
    MetaOptNet~{\citep{Lee2019Meta}}  &ResNet-12&CVPR'19& 62.64 \(\pm\) 0.35 & 78.63 \(\pm\) 0.68 & 65.99 \(\pm\) 0.72 & 81.56 \(\pm\) 0.53 \\
    ProtoMAML~{\cite{triantafillou2019meta}} &ResNet-12&ICLR'20& 64.12 \(\pm\) 0.20 & 81.24 \(\pm\) 0.20 & 68.46 \(\pm\) 0.23 & 84.67 \(\pm\) 0.16 \\

    DSN-MR~{\cite{Simon2020Adaptive}}   &ResNet-12&CVPR'20& 64.60 \(\pm\) 0.72 & 79.51 \(\pm\) 0.50 & 67.39 \(\pm\) 0.82 & 82.85 \(\pm\) 0.56 \\
    Meta-Baseline~\cite{chen2021meta} &ResNet-12&ICCV’21&  63.17 \(\pm\) 0.23 &79.26 \(\pm\) 0.17 &68.62 \(\pm\) 0.27 &83.29 \(\pm\) 0.18\\
    Unicorn-MAML~\cite{ye2021train}&ResNet-12&ICLR’22& 65.17 \(\pm\) 0.20 &84.30 \(\pm\) 0.14 &69.24 \(\pm\) 0.20 &86.06 \(\pm\) 0.16\\
    Meta-AdaM~\cite{{sun2024meta}} &ResNet-12&NeurIPS'23& 59.89 \(\pm\) 0.49 & 77.92 \(\pm\) 0.43 & 65.31 \(\pm\) 0.48 &85.24 \(\pm\) 0.35\\
    \midrule
    MSD  &ResNet-12&OURS& {\bf 65.41 \(\pm\) 0.47} & {\bf 84.88 \(\pm\) 0.29} & {\bf 69.73 \(\pm\) 0.48} & {\bf 86.25 \(\pm\) 0.29} \\
    \bottomrule
    \end{tabular}
\end{table}

\section{Experiment}

\subsection{Experiment Setting}

\textbf{Datasets.} Our methodology was primarily evaluated on two benchmark datasets: {\it Mini}ImageNet\cite{vinyals2016matching} and Tiered-ImageNet\cite{ren2018meta}, both widely used for few-shot learning assessments.

The {\it Mini}ImageNet dataset comprises 100 classes, each containing 600 samples. Samples are color images with a resolution of $84 \times 84$ pixels. Following prior work, we devide the 100 classes into training, validation, and test sets, containing 64, 16, and 20 classes, respectively. The Tiered-ImageNet dataset encompasses 608 classes with a total of 779,165 images. These fine-grained classes are categorized into 34 higher-level classes. In alignment with previous studies, we divided these higher-level classes into training, validation, and test sets, comprising 20, 6, and 8 higher-level classes, respectively. Tiered-ImageNet is designed to consider class similarity when segmenting the dataset, ensuring a significant distributional difference between training and test data.

\textbf{Backbone Model.} For our model evaluation, following\cite{Lee2019Meta}, we employed a ResNet-12\cite{he2016deep} architecture, noted for its broader widths and Dropblock modules as introduced by \cite{ghiasi2018dropblock}. This backbone is broadly used across numerous few-shot learning algorithms. Additionally, we follow the original MAML approach, utilizing a 4-layer convolutional neural network(Conv4)\cite{vinyals2016matching}. Following the recent practice\cite{ye2020few,qiao2018few,rusu2018meta}, The models' weights are pre-trained on the entire meta-training set to initialize.

\begin{table}[htbp]
  \centering
  
  \caption{\textbf{5way1shot and 5way5shot classification accuracy} in strongly augmented few-shot classification task and 95\% confidence interval on {\it Mini}ImageNet and {\it Tiered}ImageNet (over 2000 tasks), using ResNet-12 as the backbone.
  }
  \label{aug_mini}
  \vspace{-10pt}
  \tabcolsep 3pt
    \begin{tabular}{|c|c|cc|cc|}
    \addlinespace
    \toprule
    \multicolumn{ 2}{|c}{} & \multicolumn{ 2}{|c}{{\it Mini}ImageNet} & \multicolumn{ 2}{|c|}{{\it Tiered}ImageNet} \\
    \midrule
    \multicolumn{ 1}{|c|}{Methods} &Backbone  & 1-Shot & 5-Shot & 1-Shot & 5-Shot \\
    \midrule
    MAML  &ResNet-12& {49.94 $\pm$ 0.43} & {73.46 $\pm$ 0.36} & {46.35 $\pm$ 0.47} & {71.28 $\pm$ 0.42} \\
    MSD + MAML   &ResNet-12& {\bf 57.31 $\pm$ 0.44} & {\bf 78.32 $\pm$ 0.33} & {\bf 52.88 $\pm$ 0.44} & {\bf 75.06 $\pm$ 0.43} \\
    \midrule
    Unicorn-MAML   &ResNet-12 & {50.57 $\pm$ 0.43} & {73.68 $\pm$ 0.35} & {46.19 $\pm$ 0.46} & {70.35 $\pm$ 0.43} \\
    MSD + Unicorn-MAML  &ResNet-12& {\bf 57.75 $\pm$ 0.44} & {\bf 77.25 $\pm$ 0.33} & {\bf 54.80 $\pm$ 0.49} & {\bf 74.24 $\pm$ 0.41} \\
    \bottomrule
    \end{tabular}
\end{table}
\begin{table}[htbp]
  \centering
  
  \caption{\textbf{5way1shot and 5way5shot classification accuracy} in augmented few-shot classification task and 95\% confidence interval on {\it Mini}ImageNet and {\it Tiered}ImageNet (over 2000 tasks), using Conv4 as the backbone.the terms “strong” and “weak” denote the varying levels of augmentation applied to the support data in the meta-test phase.
  }
  \label{aug_conv}
  \vspace{-10pt}
  \tabcolsep 3pt
    \begin{tabular}{|c|c|cc|cc|}
    \addlinespace
    \toprule
    \multicolumn{ 2}{|c}{} & \multicolumn{ 2}{|c}{{\it Mini}ImageNet(Strong)} & \multicolumn{ 2}{|c|}{{\it Mini}ImageNet(Weak)} \\
    \midrule
    \multicolumn{ 1}{|c|}{Methods}  &Backbone& 1-Shot & 5-Shot & 1-Shot & 5-Shot \\
    \midrule
    MAML   &Conv4& {28.13 $\pm$ 0.29} & {37.77 $\pm$ 0.31}  & {35.89 $\pm$ 0.35}  & {49.54 $\pm$ 0.36}  \\
    MSD + MAML   &Conv4& {\bf 30.64 $\pm$ 0.30} & {\bf 40.79 $\pm$ 0.33} & {\bf 37.11 $\pm$ 0.37} & {\bf 50.38 $\pm$ 0.37} \\
    \midrule
    Unicorn-MAML   &Conv4& {29.26 $\pm$ 0.30} & {40.58 $\pm$ 0.33}  & {36.07 $\pm$0.36}  & {51.43 $\pm$ 0.37} \\
    MSD + Unicorn-MAML  &Conv4&  {\bf 31.37 $\pm$ 0.32} & {\bf 42.59 $\pm$ 0.33} & {\bf 38.94 $\pm$ 0.38} & {\bf 54.11 $\pm$ 0.37}  \\
    \bottomrule
    \end{tabular}
\end{table}

\subsection{Results}
\subsubsection{Standard Few-shot Learning Problems.}
The results in Tab.\ref{standard_fsl} demonstrate the performance of MSD and several mainstream few-shot algorithms on few-shot tasks. MSD exhibits a significant improvement over MAML in traditional few-shot tasks. On {\it Mini}ImageNet, our method achieved an increase of 0.99\% in 5way1shot and 1.44\% in 5way5shot tasks, respectively. On Tiered ImageNet, the improvements for 5way1shot and 5way5shot tasks were 4.11\% and 1.61\%, respectively. MSD shows excellent effectiveness in few-shot tasks, with better performance compared to the recent meta-learning algorithms and MAML's variants.

\subsubsection{Augmented Few-shot Learning Problems.}
To further investigate the enhancement of model precision learning capabilities through MSD, we employed augmented tasks to test the model. Specifically, during the meta-test phase, we augmented the support data to fine-tune the model and then classified the query data using the updated model. We report the accuracy of model classification and the consistency of knowledge learned across different methods. Conv4 and Resnet12 were utilized to validate that MSD can improve the precision learning abilities of models of varying scales.

\begin{table}[htbp]
  \centering
   \vskip -5pt
  
  \caption{\textbf{5way1shot and 5way5shot consistency of learned knowledge} in strong augmented few-shot classification task on {\it Mini}ImageNet and {\it Tiered}ImageNet (over 2000 tasks), using ResNet-12 as the backbone.
  }
  \label{aug_mini_sim}
  \vspace{-10pt}
  \tabcolsep 3pt
    \begin{tabular}{|c|c|cc|cc|}
    \addlinespace
    \toprule
    \multicolumn{2}{|c}{} & \multicolumn{ 2}{|c}{{\it Mini}ImageNet} & \multicolumn{ 2}{|c|}{{\it Tiered}ImageNet} \\
    \midrule
    \multicolumn{ 1}{|c|}{Methods}& \multicolumn{ 1}{|c|}{Backbone}  & 1-Shot & 5-Shot & 1-Shot & 5-Shot \\
    \midrule

    MAML  &ResNet-12 & {85.88} & {94.03} & {86.79 } & {93.87} \\
    MSD + MAML &ResNet-12  & {\bf 98.58} & {\bf 99.00} & {\bf 99.63} & {\bf 99.80} \\
    \midrule
    Unicorn-MAML & ResNet-12  & {87.55} & {94.60} & {87.81} & {95.41} \\
    MSD + Unicorn-MAML & ResNet-12 & {\bf 99.91} & {\bf 99.92} & {\bf 99.94} & {\bf 99.96} \\
    \bottomrule
    \end{tabular}
\end{table}

\textbf{Augmented few-shot accuracy.}
Tab.\ref{aug_conv} presents the performance of Conv4 on the {\it Mini}ImageNet dataset under varying levels of augmentation. MSD has an approximate 2\% increase in classification accuracy on query data, irrespective of whether the perturbations are weak or strong. Tab.\ref{aug_mini} demonstrates the performance of ResNet-12 under strong augmentation on both {\it Mini}ImageNet and Tiered ImageNet datasets. It is evident that MSD confers greater improvements on models with larger capacities. Specifically, MSD contributes to an approximate 7\% increase in accuracy for 5way1shot tasks and about a 4\% increase for 5way5shot tasks.

\textbf{Consitency of learned knowledge.}
Tab.\ref{aug_mini_sim} presents the consistency of knowledge acquired by the model variants for the same support data, as quantified by the cosine similarity among the outputs of different model versions for the same query data, as shown in Eq.\ref{eq9}. It is observed that both MAML and its variant, MAML-Unicorn, tend to learn biased knowledge in the 5way1shotscenario, resulting in a lower consistency of approximately 86\%. In the 5way5shot scenario, the models exhibit reduced hypothesis redundancy, thereby increasing the consistency of learned knowledge to approximately 94\%. Our proposed Meta Self-Distillation (MSD) approach significantly enhances the model’s extraction of precise knowledge, achieving around 99\% consistency in knowledge across both datasets for 5way1shot and 5way5shot problems.

\begin{figure}[tbp]
	\centering
	\includegraphics[width=1.0\columnwidth]{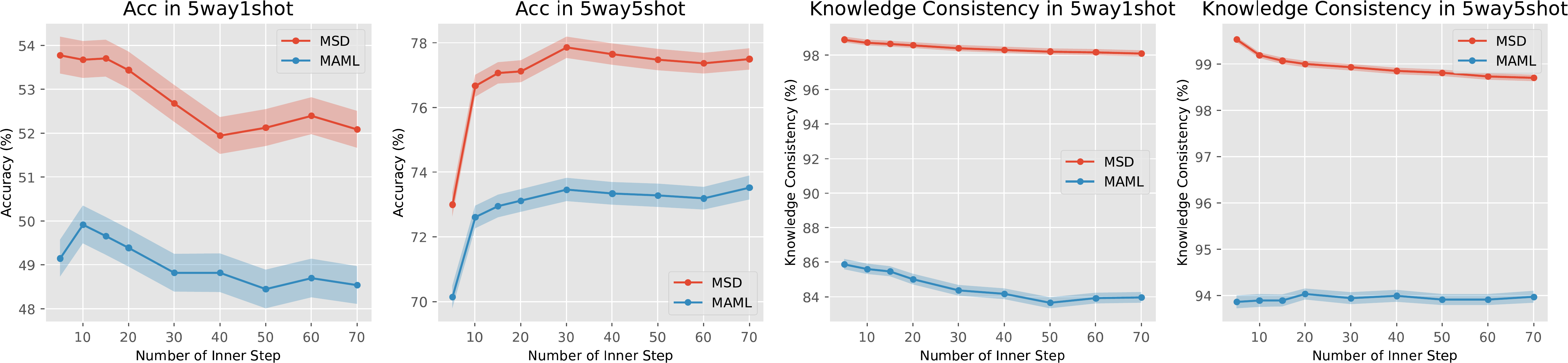}
	\caption{\textbf{The 5way1shot and 5way5shot classification accuracy and the consistency of learned knowledge with different numbers of inner steps} with 95\% confidence interval, averaged over 2000 tasks}
    \label{innerstep}
\end{figure}

\begin{table}[h]
\caption{\textbf{Ablation study on {\it Mini}ImageNet.} All models are trained on the full training set of {\it Mini}ImageNet. }.
\label{ablation}
\centering
\begin{tabular}{cccccc}
\Cline{0.6pt}{1-5}
\\[-0.81em]
\multirow{2}{*}{Second Order}  & \multirow{2}{*}{Augmentation} &
\multirow{2}{*}{$\mathcal L_{\text{Knowledge-Consistency}}$} &
\multicolumn{2}{c}{{\it Mini}ImageNet}  \\
& & & 
1-shot & 5-shot \\ \hline 

\Cline{0.6pt}{1-5}
\\[-8pt] 

\ding{55} & \ding{51} & \ding{51} & 64.51\scriptsize$\pm$ 0.48


& 84.45
\scriptsize$\pm$ 0.28
 \\
\ding{51} & \ding{55} &\ding{55} & 64.43\scriptsize$\pm$ 0.46

& 83.90
\scriptsize$\pm$ 0.29

\\
\ding{51} &\ding{51} & \ding{55} & 64.31\scriptsize$\pm$ 0.48
& 84.14
\scriptsize$\pm$ 0.28

\\

\ding{51}  &\ding{51} & \ding{51} &  \bf 65.41\scriptsize$\pm$ 0.47
& \bf 84.88 \scriptsize$\pm$ 0.29
\\

\Cline{0.6pt}{1-5}
\end{tabular}
\end{table}

\subsection{Ablation Study}

\textbf{The impact of each component.}
Tab.\ref{ablation} demonstrates three principal factors influencing model performance during the optimization process of MSD: the use of second-order derivatives, data augmentation on support data, and the employment of MSD’s knowledge consistency loss. It is observed that MSD optimized with first-order derivatives can achieve successful optimization, albeit at the expense of little few-shot learning performance. The efficacy of MSD’s knowledge consistency loss is contingent upon data augmentation applied to support data during the meta-training phase; hence, in the absence of data augmentation, the knowledge consistency loss is rendered ineffective. Rows two and three of the table indicate a marginal performance enhancement attributed to data augmentation, though the improvement is not substantial. The main contribution to the MSD’s performance enhancement is derived from the knowledge consistency loss.

\textbf{The impact of the inner step.}
Concurrently, we further investigated the impact of different inner steps during the meta-test phase on the model’s few-shot classification accuracy and precise learning capabilities.Fig.\ref{innerstep} illustrates the impact of the number of inner steps during the meta-test phase on the performance of the MSD algorithm. The results indicate that for any given number of inner steps, the models trained using MSD consistently outperformed those trained with MAML. Specifically, in the 5way1shot and 5way5shot tasks, MSD achieved an accuracy of approximately 7\% and 4\% higher than MAML, respectively. Concerning the consistency of the knowledge learned, there was a trend of decreasing consistency for both MAML and MSD as the number of inner steps increased. This suggests that an excessive number of inner steps during the meta-test phase may lead to the model learning shortcut features. However, MSD maintained approximately 99\% consistency in the knowledge learned for both the 5way1shot and 5way5shot tasks, significantly surpassing the performance of MAML.

\begin{figure}[tbp]
	\centering
	\includegraphics[width=1.0\columnwidth]{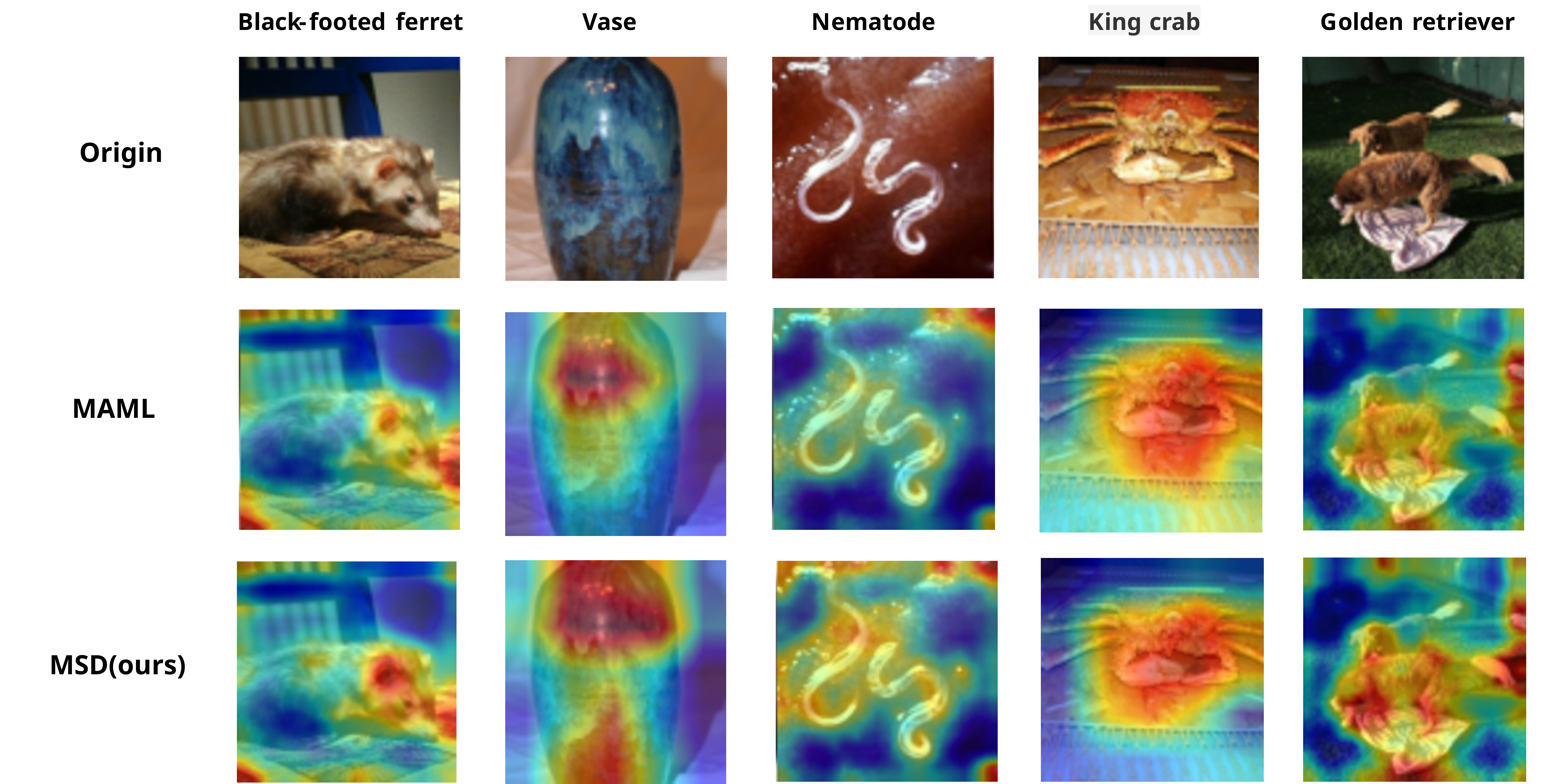}
	\caption{\textbf{The results of the visual analysis} on the test set of {\it Mini}ImageNet with MAML and MSD.}
    \label{vis}
\end{figure}

\subsection{Visualization}

To further analyze the MSD on the learning capabilities of models, we visualized the models updated by augmented data as shown in Fig.\ref{vis}. Specifically, during the meta-test phase, we visualized models trained with Model-Agnostic Meta-Learning (MAML) and MSD. The model was first fine-tuned using augmented support data, with the number of inner steps set to 20. Then, query data was employed as the evaluate data. Grad-CAM++\cite{chattopadhay2018grad} was utilized to visualize the critical regions that the models focused on for understanding the query data. The visualizations reveal that the model trained with MAML tends to allocate more attention to the surrounding environment, potentially prioritizing it over the classified objects, while the model trained with MSD focuses more on the objects used for classification.

\section{Conclusion}
In this work, we propose the objective of enabling models to learn with precision, aspiring for the model to acquire category-related invariant features from the training data while diminishing attention to biased or shortcut features. Building on this foundation, we introduce a meta self-distillation optimization framework. This framework updates the model variants by utilizing different variants of support data in an inner loop. The initial parameters’ precision learning capability is then assessed based on the consistency of their outputs for the same query data. Experiments demonstrate the effectiveness of our algorithm on few-shot tasks, and notably, on perturbed few-shot tasks, MSD significantly enhances the performance of algorithms such as MAML.

The ability for models to learn with greater precision is of paramount importance. We believe our proposed algorithm represents a step forward in enhancing models’ ability to learn more precisely. Future research could extend such a framework to the domain of self-supervised learning and apply it to larger-scale models.




\begin{thebibliography}{37}
\providecommand{\natexlab}[1]{#1}
\providecommand{\url}[1]{\texttt{#1}}
\expandafter\ifx\csname urlstyle\endcsname\relax
  \providecommand{\doi}[1]{doi: #1}\else
  \providecommand{\doi}{doi: \begingroup \urlstyle{rm}\Url}\fi

\bibitem[Allen-Zhu and Li(2020)]{allen2020towards}
Z.~Allen-Zhu and Y.~Li.
\newblock Towards understanding ensemble, knowledge distillation and self-distillation in deep learning.
\newblock \emph{arXiv preprint arXiv:2012.09816}, 2020.

\bibitem[Antoniou et~al.(2018)Antoniou, Edwards, and Storkey]{antoniou2018train}
A.~Antoniou, H.~Edwards, and A.~Storkey.
\newblock How to train your maml.
\newblock In \emph{International conference on learning representations}, 2018.

\bibitem[Caron et~al.(2020)Caron, Misra, Mairal, Goyal, Bojanowski, and Joulin]{caron2020unsupervised}
M.~Caron, I.~Misra, J.~Mairal, P.~Goyal, P.~Bojanowski, and A.~Joulin.
\newblock Unsupervised learning of visual features by contrasting cluster assignments.
\newblock \emph{Advances in neural information processing systems}, 33:\penalty0 9912--9924, 2020.

\bibitem[Chattopadhay et~al.(2018)Chattopadhay, Sarkar, Howlader, and Balasubramanian]{chattopadhay2018grad}
A.~Chattopadhay, A.~Sarkar, P.~Howlader, and V.~N. Balasubramanian.
\newblock Grad-cam++: Generalized gradient-based visual explanations for deep convolutional networks.
\newblock In \emph{2018 IEEE winter conference on applications of computer vision (WACV)}, pages 839--847. IEEE, 2018.

\bibitem[Chen et~al.(2020)Chen, Kornblith, Norouzi, and Hinton]{chen2020simple}
T.~Chen, S.~Kornblith, M.~Norouzi, and G.~Hinton.
\newblock A simple framework for contrastive learning of visual representations.
\newblock In \emph{International conference on machine learning}, pages 1597--1607. PMLR, 2020.

\bibitem[Chen et~al.(2019)Chen, Liu, Kira, Wang, and Huang]{chen2019closer}
W.-Y. Chen, Y.-C. Liu, Z.~Kira, Y.-C.~F. Wang, and J.-B. Huang.
\newblock A closer look at few-shot classification.
\newblock \emph{arXiv preprint arXiv:1904.04232}, 2019.

\bibitem[Chen and He(2021)]{chen2021exploring}
X.~Chen and K.~He.
\newblock Exploring simple siamese representation learning.
\newblock In \emph{Proceedings of the IEEE/CVF conference on computer vision and pattern recognition}, pages 15750--15758, 2021.

\bibitem[Chen et~al.(2021)Chen, Liu, Xu, Darrell, and Wang]{chen2021meta}
Y.~Chen, Z.~Liu, H.~Xu, T.~Darrell, and X.~Wang.
\newblock Meta-baseline: Exploring simple meta-learning for few-shot learning.
\newblock In \emph{Proceedings of the IEEE/CVF international conference on computer vision}, pages 9062--9071, 2021.

\bibitem[Finn et~al.(2017)Finn, Abbeel, and Levine]{finn2017model}
C.~Finn, P.~Abbeel, and S.~Levine.
\newblock Model-agnostic meta-learning for fast adaptation of deep networks.
\newblock In \emph{International conference on machine learning}, pages 1126--1135. PMLR, 2017.

\bibitem[Ghiasi et~al.(2018)Ghiasi, Lin, and Le]{ghiasi2018dropblock}
G.~Ghiasi, T.-Y. Lin, and Q.~V. Le.
\newblock Dropblock: A regularization method for convolutional networks.
\newblock \emph{Advances in neural information processing systems}, 31, 2018.

\bibitem[Grill et~al.(2020)Grill, Strub, Altch{\'e}, Tallec, Richemond, Buchatskaya, Doersch, Avila~Pires, Guo, Gheshlaghi~Azar, et~al.]{grill2020bootstrap}
J.-B. Grill, F.~Strub, F.~Altch{\'e}, C.~Tallec, P.~Richemond, E.~Buchatskaya, C.~Doersch, B.~Avila~Pires, Z.~Guo, M.~Gheshlaghi~Azar, et~al.
\newblock Bootstrap your own latent-a new approach to self-supervised learning.
\newblock \emph{Advances in neural information processing systems}, 33:\penalty0 21271--21284, 2020.

\bibitem[He et~al.(2016)He, Zhang, Ren, and Sun]{he2016deep}
K.~He, X.~Zhang, S.~Ren, and J.~Sun.
\newblock Deep residual learning for image recognition.
\newblock In \emph{Proceedings of the IEEE conference on computer vision and pattern recognition}, pages 770--778, 2016.

\bibitem[He et~al.(2020)He, Fan, Wu, Xie, and Girshick]{he2020momentum}
K.~He, H.~Fan, Y.~Wu, S.~Xie, and R.~Girshick.
\newblock Momentum contrast for unsupervised visual representation learning.
\newblock In \emph{Proceedings of the IEEE/CVF conference on computer vision and pattern recognition}, pages 9729--9738, 2020.

\bibitem[Hinton et~al.(2015)Hinton, Vinyals, and Dean]{hinton2015distilling}
G.~Hinton, O.~Vinyals, and J.~Dean.
\newblock Distilling the knowledge in a neural network.
\newblock \emph{arXiv preprint arXiv:1503.02531}, 2015.

\bibitem[Jamal and Qi(2019)]{jamal2019task}
M.~A. Jamal and G.-J. Qi.
\newblock Task agnostic meta-learning for few-shot learning.
\newblock In \emph{Proceedings of the IEEE/CVF conference on computer vision and pattern recognition}, pages 11719--11727, 2019.

\bibitem[Kao et~al.(2021)Kao, Chiu, and Chen]{kao2021maml}
C.-H. Kao, W.-C. Chiu, and P.-Y. Chen.
\newblock Maml is a noisy contrastive learner in classification.
\newblock \emph{arXiv preprint arXiv:2106.15367}, 2021.

\bibitem[Lake and Baroni(2023)]{lake2023human}
B.~M. Lake and M.~Baroni.
\newblock Human-like systematic generalization through a meta-learning neural network.
\newblock \emph{Nature}, 623\penalty0 (7985):\penalty0 115--121, 2023.

\bibitem[Le et~al.(2021)Le, Nguyen, Nguyen, Tran, Nguyen, and Hua]{le2021poodle}
D.~Le, K.~D. Nguyen, K.~Nguyen, Q.-H. Tran, R.~Nguyen, and B.-S. Hua.
\newblock Poodle: Improving few-shot learning via penalizing out-of-distribution samples.
\newblock \emph{Advances in Neural Information Processing Systems}, 34:\penalty0 23942--23955, 2021.

\bibitem[Lee et~al.(2019)Lee, Maji, Ravichandran, and Soatto]{Lee2019Meta}
K.~Lee, S.~Maji, A.~Ravichandran, and S.~Soatto.
\newblock Meta-learning with differentiable convex optimization.
\newblock In \emph{Proceedings of the IEEE/CVF conference on computer vision and pattern recognition}, pages 10657--10665, 2019.

\bibitem[Li et~al.(2022)Li, Qiang, Zheng, Su, and Xiong]{li2022metaug}
J.~Li, W.~Qiang, C.~Zheng, B.~Su, and H.~Xiong.
\newblock Metaug: Contrastive learning via meta feature augmentation.
\newblock In \emph{International Conference on Machine Learning}, pages 12964--12978. PMLR, 2022.

\bibitem[Ni et~al.(2021)Ni, Shu, Souri, Goldblum, and Goldstein]{ni2021close}
R.~Ni, M.~Shu, H.~Souri, M.~Goldblum, and T.~Goldstein.
\newblock The close relationship between contrastive learning and meta-learning.
\newblock In \emph{International conference on learning representations}, 2021.

\bibitem[Nichol et~al.(2018)Nichol, Achiam, and Schulman]{nichol2018first}
A.~Nichol, J.~Achiam, and J.~Schulman.
\newblock On first-order meta-learning algorithms.
\newblock \emph{arXiv preprint arXiv:1803.02999}, 2018.

\bibitem[Qiao et~al.(2018)Qiao, Liu, Shen, and Yuille]{qiao2018few}
S.~Qiao, C.~Liu, W.~Shen, and A.~L. Yuille.
\newblock Few-shot image recognition by predicting parameters from activations.
\newblock In \emph{Proceedings of the IEEE conference on computer vision and pattern recognition}, pages 7229--7238, 2018.

\bibitem[Raghu et~al.(2019)Raghu, Raghu, Bengio, and Vinyals]{raghu2019rapid}
A.~Raghu, M.~Raghu, S.~Bengio, and O.~Vinyals.
\newblock Rapid learning or feature reuse? towards understanding the effectiveness of maml.
\newblock \emph{arXiv preprint arXiv:1909.09157}, 2019.

\bibitem[Ren et~al.(2018)Ren, Triantafillou, Ravi, Snell, Swersky, Tenenbaum, Larochelle, and Zemel]{ren2018meta}
M.~Ren, E.~Triantafillou, S.~Ravi, J.~Snell, K.~Swersky, J.~B. Tenenbaum, H.~Larochelle, and R.~S. Zemel.
\newblock Meta-learning for semi-supervised few-shot classification.
\newblock \emph{arXiv preprint arXiv:1803.00676}, 2018.

\bibitem[Rusu et~al.(2018)Rusu, Rao, Sygnowski, Vinyals, Pascanu, Osindero, and Hadsell]{rusu2018meta}
A.~A. Rusu, D.~Rao, J.~Sygnowski, O.~Vinyals, R.~Pascanu, S.~Osindero, and R.~Hadsell.
\newblock Meta-learning with latent embedding optimization.
\newblock \emph{arXiv preprint arXiv:1807.05960}, 2018.

\bibitem[Simon et~al.(2020)Simon, Koniusz, Nock, and Harandi]{Simon2020Adaptive}
C.~Simon, P.~Koniusz, R.~Nock, and M.~Harandi.
\newblock Adaptive subspaces for few-shot learning.
\newblock In \emph{Proceedings of the IEEE/CVF conference on computer vision and pattern recognition}, pages 4136--4145, 2020.

\bibitem[Snell et~al.(2017)Snell, Swersky, and Zemel]{snell2017prototypical}
J.~Snell, K.~Swersky, and R.~Zemel.
\newblock Prototypical networks for few-shot learning.
\newblock \emph{Advances in neural information processing systems}, 30, 2017.

\bibitem[Sun et~al.(2020)Sun, Liu, Chen, Chua, and Schiele]{sun2020meta}
Q.~Sun, Y.~Liu, Z.~Chen, T.-S. Chua, and B.~Schiele.
\newblock Meta-transfer learning through hard tasks.
\newblock \emph{IEEE Transactions on Pattern Analysis and Machine Intelligence}, 44\penalty0 (3):\penalty0 1443--1456, 2020.

\bibitem[Sun and Gao(2024)]{sun2024meta}
S.~Sun and H.~Gao.
\newblock Meta-adam: An meta-learned adaptive optimizer with momentum for few-shot learning.
\newblock \emph{Advances in Neural Information Processing Systems}, 36, 2024.

\bibitem[Sung et~al.(2018)Sung, Yang, Zhang, Xiang, Torr, and Hospedales]{sung2018learning}
F.~Sung, Y.~Yang, L.~Zhang, T.~Xiang, P.~H. Torr, and T.~M. Hospedales.
\newblock Learning to compare: Relation network for few-shot learning.
\newblock In \emph{Proceedings of the IEEE conference on computer vision and pattern recognition}, pages 1199--1208, 2018.

\bibitem[Triantafillou et~al.(2019)Triantafillou, Zhu, Dumoulin, Lamblin, Evci, Xu, Goroshin, Gelada, Swersky, Manzagol, et~al.]{triantafillou2019meta}
E.~Triantafillou, T.~Zhu, V.~Dumoulin, P.~Lamblin, U.~Evci, K.~Xu, R.~Goroshin, C.~Gelada, K.~Swersky, P.-A. Manzagol, et~al.
\newblock Meta-dataset: A dataset of datasets for learning to learn from few examples.
\newblock \emph{arXiv preprint arXiv:1903.03096}, 2019.

\bibitem[Vinyals et~al.(2016)Vinyals, Blundell, Lillicrap, Wierstra, et~al.]{vinyals2016matching}
O.~Vinyals, C.~Blundell, T.~Lillicrap, D.~Wierstra, et~al.
\newblock Matching networks for one shot learning.
\newblock \emph{Advances in neural information processing systems}, 29, 2016.

\bibitem[Wang et~al.(2020)Wang, Yao, Kwok, and Ni]{wang2020generalizing}
Y.~Wang, Q.~Yao, J.~T. Kwok, and L.~M. Ni.
\newblock Generalizing from a few examples: A survey on few-shot learning.
\newblock \emph{ACM computing surveys (csur)}, 53\penalty0 (3):\penalty0 1--34, 2020.

\bibitem[Ye and Chao(2021)]{ye2021train}
H.-J. Ye and W.-L. Chao.
\newblock How to train your maml to excel in few-shot classification.
\newblock \emph{arXiv preprint arXiv:2106.16245}, 2021.

\bibitem[Ye et~al.(2020)Ye, Hu, Zhan, and Sha]{ye2020few}
H.-J. Ye, H.~Hu, D.-C. Zhan, and F.~Sha.
\newblock Few-shot learning via embedding adaptation with set-to-set functions.
\newblock In \emph{Proceedings of the IEEE/CVF conference on computer vision and pattern recognition}, pages 8808--8817, 2020.

\bibitem[Zhou et~al.(2023)Zhou, Wang, Zhang, Wei, and Zhang]{zhou2023revisiting}
F.~Zhou, P.~Wang, L.~Zhang, W.~Wei, and Y.~Zhang.
\newblock Revisiting prototypical network for cross domain few-shot learning.
\newblock In \emph{Proceedings of the IEEE/CVF Conference on Computer Vision and Pattern Recognition}, pages 20061--20070, 2023.

\end{thebibliography}


\appendix

\section{Appendix / supplemental material}
\subsection{Limitations}
\label{limitation}
In this paper, we only discuss the standard and augmented few-shot scenarios, while the method proposed
can be used in many other existing fields. Also, due to the costly second-order derivative involved, it is computation/memory expensive to apply the method to a larger model.
\subsection{Hyperparameters And Code Environment Of Experiment}
\label{Reproducibility}
\textbf{Hyperparameters.}

The hyperparameters has shown in the Tab.\ref{para}Tab.\ref{strong aug}Tab.\ref{weak aug}

\textbf{Calculate resources and Environment.}
Our experiment is conducted on NVIDIA A800 80GB PCIe. The software environment consisted of Python version 3.10.14, and PyTorch version 2.3.0, with CUDA toolkit 12.1

\subsection{Broader Impact}

\label{Broader Impact}
The method we proposed enhances the model’s ability to learn more precisely. In domains such as healthcare, this allows for good performance with only a small amount of data, which brings positive social impact.

\begin{table}[h]
\centering
\caption{Experimental Setup}
\label{para}
\begin{tabular}{|l|l|}
\hline
\textbf{Parameter} & \textbf{Value} \\
\hline
Task Batch Size & 2 \\
Inner Step Count & 20 \\
Inner Loop Learning Rate & 0.05 \\
Outer Loop Learning Rate & 0.001 \\
Query Data Points & 15 \\
Outer Loop Learning Rate Decay & 1/10 every 10 epochs \\
Coefficient \(\alpha\) (Eq.\ref{total loss}) & 1 \\
\hline
\end{tabular}

\end{table}

\begin{table}[h]
\vspace{-10pt}
\centering
\caption{Augmentations for Strong-Augmented Few-Shot Scenario}
\label{strong aug}
\begin{tabular}{|l|l|l|}
\hline
\textbf{Augmentation} & \textbf{Parameters} & \textbf{Probability} \\
\hline
Random Resize & (scale: 0.5–1) & - \\
Color Jitter & (0.8, 0.8, 0.8, 0.2) & 0.8 \\
Grayscale Conversion & - & 0.2 \\
Gaussian Blur & Expectation: 0.1, Variance: 2 & 0.5 \\
Random Horizontal Flip & - & 0.5 \\
\hline
\end{tabular}

\end{table}

\begin{table}[h]
\label{weak aug}
\centering
\caption{Augmentations for Weak-Augmented Few-Shot Scenario}
\begin{tabular}{|l|l|l|}
\hline
\textbf{Augmentation} & \textbf{Parameters} & \textbf{Probability} \\
\hline
Center Crop & 84 \(\times\) 84 & - \\
Color Jitter & (0.4, 0.4, 0.4, 0.1) & 0.8 \\
Grayscale Conversion & - & 0.2 \\
Gaussian Blur & Expectation: 0, Variance: 1 & 0.5 \\
Random Horizontal Flip & - & 0.5 \\
\hline
\end{tabular}

\end{table}

\end{document}